\pdfoutput=1

\documentclass[11pt]{article}
\usepackage[dvipsnames]{xcolor}
\usepackage{authblk}
\usepackage{naacl2021}

\usepackage{times}
\usepackage{latexsym}

\usepackage[T1]{fontenc}

\usepackage[utf8]{inputenc}

\usepackage{microtype}

\usepackage{amsmath}

\usepackage{dsfont}
\usepackage{graphicx}
\usepackage{caption}

\DeclareMathOperator*{\argmax}{argmax}

\usepackage{array}
\newcolumntype{P}[1]{>{\centering\arraybackslash}p{#1}}
\usepackage[ruled,vlined]{algorithm2e}
\title{Seeing is Knowing! Fact-based Visual Question Answering using Knowledge Graph Embeddings}

\author[1]{Kiran Ramnath}
\author[1]{Mark Hasegawa-Johnson}
\affil[1]{University of Illinois, Urbana-Champaign}
\affil[1]{\texttt{\{kiranr2, jhasegaw\}@illinois.edu}}

\begin{document}
\maketitle
\begin{abstract}
Fact-based Visual Question Answering (FVQA), a challenging variant of VQA, requires a QA-system to include facts from a diverse knowledge graph (KG) in its reasoning process to produce an answer. Large KGs, especially common-sense KGs, are known to be incomplete, i.e., not all non-existent facts are always incorrect. Therefore, being able to reason over incomplete KGs for QA is a critical requirement in real-world applications that has not been addressed extensively in the literature. We develop a novel QA architecture that allows us to reason over incomplete KGs, something current FVQA state-of-the-art (SOTA) approaches lack due to their critical reliance on fact retrieval. We use KG Embeddings, a technique widely used for KG completion, for the downstream task of FVQA. We also employ a new image representation technique we call ‘Image-as-Knowledge’ to enable this capability, alongside a simple one-step CoAttention mechanism to attend to text and image during QA. Our FVQA architecture is faster during inference time, being $O(m)$, as opposed to existing FVQA SOTA methods which are $O(N \log N)$, where $m =$ number of vertices, $N =$ number of edges $=  O(m^2)$. KG embeddings are shown to hold complementary information to word embeddings: a combination of both metrics permits performance comparable to SOTA methods in the standard answer retrieval task, and significantly better (26\% absolute) in the proposed missing-edge reasoning task.
\end{abstract}

\section{Introduction}

Multi-modal reasoning has been a prime focus in the pursuit of artificial general intelligence.
One such task is Visual Question Answering~\cite{VQA1.0,VQA2.0}, which requires the system to answer a textual question based on an image. VQA is challenging
because it requires many capabilities such as object detection, scene recognition, activity recognition in addition to language understanding and commonsense reasoning.
\begin{figure}[t]
\centering
\includegraphics[width=0.7\columnwidth]{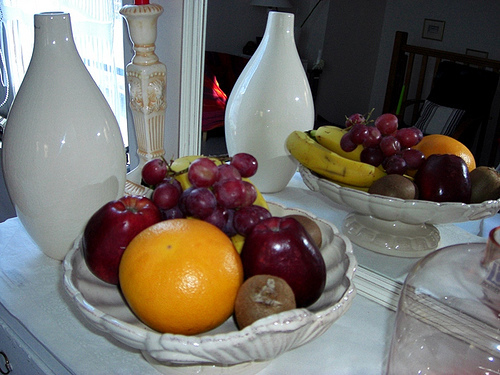} 
\centering
\caption{Example of a fact-based visual question \\
\textbf{Question} - Which object in this image is rich in potassium? \\
\textbf{Supporting fact} - Bananas are rich in potassium \\
\textbf{SPO triple} - (Bananas, \textsc{HasProperty}, rich in potassium) \\
\textbf{Answer Source} - Image \\
\textbf{Answer} - Bananas}
\label{fig:Example}
\end{figure}
~\cite{FVQA} noted that most of the questions in the VQA dataset didn't require commonsense knowledge residing outside of the image. Humans have a remarkable ability to blend in knowledge from their own prior experiences when answering a question about an image. \cite{FVQA} therefore introduced the FVQA benchmark, aiming to provide a more challenging set of questions which ensure that the answer to a given question requires some form of external knowledge, not present in the image or the question text. They provide this external information in the form of knowledge graphs, which are multi-relational graphs, storing relational representations between entities. The task of FVQA boils down to retrieving the correct entity as the answer most relevant to an image. For example, in Fig.~\ref{fig:Example} the fact triple – (Banana, \textsc{HasProperty}, rich in Potassium) is the external information required to answer the question; with the correct answer being `Banana'. 

All published methods in the literature rely on retrieving the ground-truth fact for a question, whereas a real world application will likely have questions asked about facts not present in the KG. The primary contribution of this paper is a method Seeing is Knowing (SiK) that permits FVQA to reason about common-sense facts that are absent from the knowledge graph using KG Embeddings. When used with a more general score function we define, the resultant framework works well in both complete and incomplete KG settings. KG embeddings permit us to offer two additional contributions to the SOTA in FVQA: an `image-as-knowledge'  representation of visual information, and a CoAttention method for combining visual and textual inputs.  `Image-as-knowledge' represents the image as the span of the KG embedding vectors for the entities found in it.  Representing an image with textual semantics has been attempted before \cite{Comprehension}, but not using KG embeddings; KG embeddings provide robustness to incomplete KGs by encoding information about the graph structure. CoAttention uses the words of the query to compute weighted combinations of the entity vectors, thus the answer is a vector within the span of the image entities. Two such identical models are trained separately, which are gated in order to query the entities in the KG during inference. Finally, SiK when used in a standalone fashion over incomplete KGs is more time-efficient during inference. It is $O\left(m\right)$ as it only needs to reason over existent nodes in the network. To the best of our knowledge, our work is the first one to apply KG embeddings to a VQA task.

\section{Related Work}

\subsection{Knowledge Graphs}
Knowledge graphs~\cite{Yago,DBPedia,NELL,Freebase,ERMLP} have been studied as an effective way of representing objects or concepts and their inter-relationships. Such relational representations are formally defined in the RDF (Resource Description Framework) as triples $f = (\texttt{subject}, \texttt{predicate}, \texttt{object})$, where $(\texttt{subject, object})$ are entities, $\texttt{predicate}$ is the relation connecting the two entities. Such linked representations have been found to correlate highly with how humans process cognitive information ~\cite{Cognitive}.

\subsection{KG embeddings}
Common-sense KGs extracted from web-scale texts are usually incomplete. KG embedding techniques~\cite{TransE,RotatE,NTN,RESCAL,ERMLP,ConvE} have been studied as a means to remediate incompleteness of large-scale KGs. Score-based latent feature models, which perform well on several benchmarks, learn a score mapping $\phi(h,r,t):\mathcal{E\times R \times E} \rightarrow \Re$ where $\mathcal{E}$ is the set of all entities, $\mathcal{R}$ is the set of all relation-types. $h, t \in \mathcal{E}$ are the head (\texttt{subject}) and tail (\texttt{object}), $r \in \mathcal{R}$ is the relation. $\mathcal{G} \subset \mathcal{E \times R \times E}$ is the observed set of edges, while $\mathcal{G}_o \supset \mathcal{G}$ is the unknown set of true edges. The embeddings $(h,r,t)$ are learned so that the score $\phi(.)$ is high for edges in $\mathcal{G}_o$, and low for edges in its complement.

We use Entity-Relation Multi-Layer Perceptron (ERMLP)~\cite{ERMLP} which uses an MLP to produce the score $\phi(h,r,t)$ for each fact triple. 
Distance-based models learn embeddings $h$, $r$ and $t$ in order to minimize the distance between $t$ and $f(h,r)$, for some projection function $f(\cdot)$. We explore the use of two such distance-based models (results in Appendix): TransE~\cite{TransE} and RotatE~\cite{RotatE}. TransE models $f(h,r)=h+r$. RotatE models $f(h,r)$ as a Hadamard product, $h\circ r$, where the embedding vectors $h$, $r$ and $t$ are in the complex plane and $r$ is enforced to be unit norm, so that $h\circ r$ is a rotation in the complex plane.

Knowledge-graph question answering (KGQA) is the task of answering questions using the facts in a KG.  EmbedKGQA ~\cite{EmbedKGQA} tries to reason over missing edges for text-only multi-hop questions using KG embeddings. Our task involves the visual modality in addition, as well as reasoning over a common-sense KG. 

\begin{figure*}[t]
  \centering
  \includegraphics[width=1\textwidth, height = 0.5\textwidth]{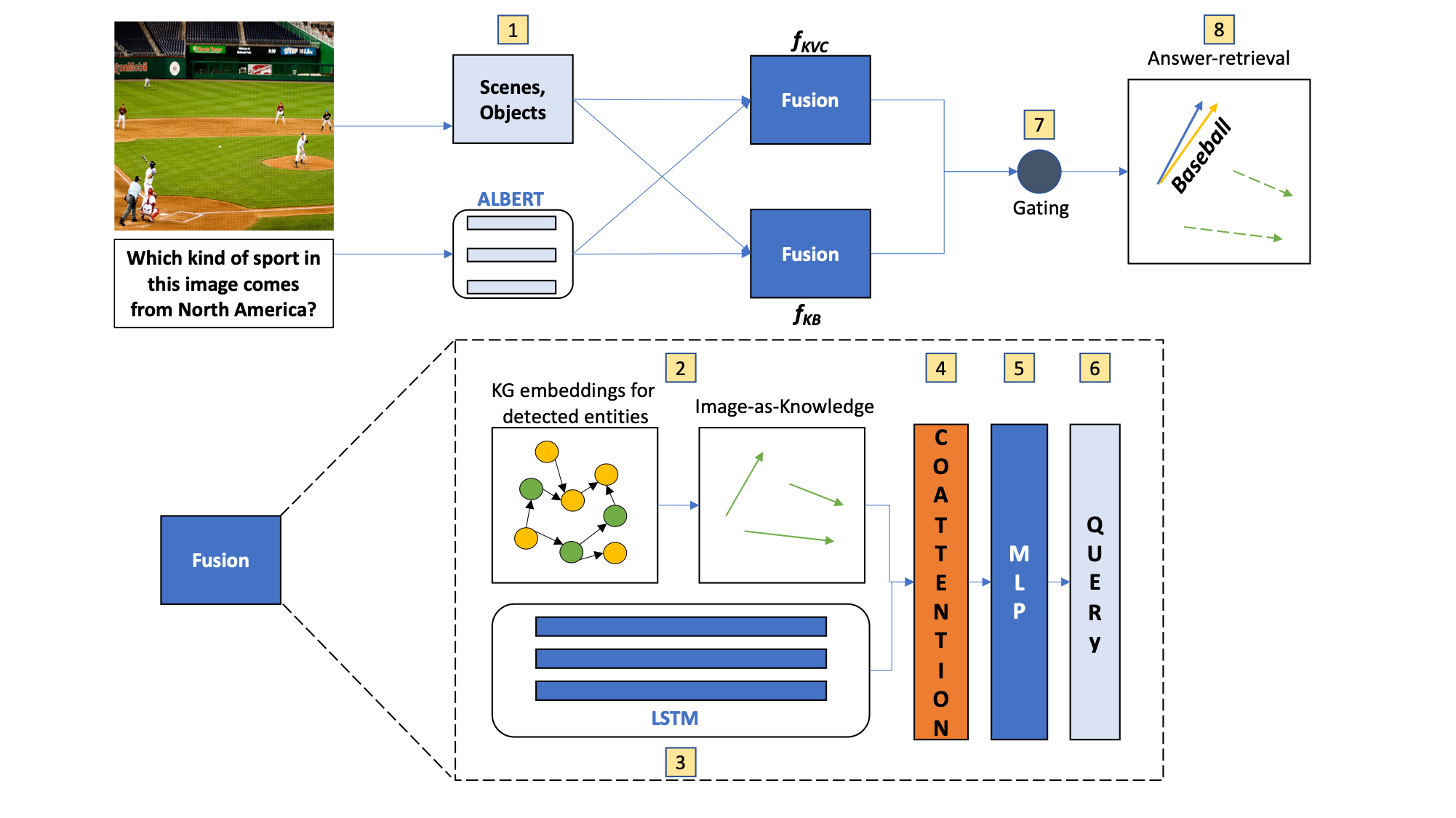}
  \caption{Seeing is Knowing architecture. (1) Scenes, objects, \& actions are detected in the image. (2) For detected entities, we retrieve their KG embedding representations. The span of these embedding vectors represents the `image-as-knowledge'. (3) Lexical semantic vectors for each word in the query are accumulated via an LSTM. (4) The joint image-question encoding is derived using a CoAttention mechanism described in Fig.~\ref{fig:CoAttention}, then (5) passed through a multi-layer perceptron, whose (6) last layer is used as a pair of queries that are (7) gated, and (8) used to retrieve the entity to answer the question. }
  \label{fig:Architecture}
\end{figure*}

\subsection{FVQA}
The fact-based visual question answering (FVQA~\cite{FVQA}) benchmark was designed using questions that obey the following condition:
for each (question,image,answer) triplet in the dataset ($(q_i,I_i,y_i)\in{\mathcal D}$), there exists exactly one supporting fact in the knowledge graph ($f_j=(h,r,t)\in{\mathcal G}$) such that the correct answer $y_i$ is either the head or the tail of $f_j$, and such that at least one of the two entities is visible in the image. 


The accompanying KG is also diverse, comprising facts from three individual KGs: Webchild ~\cite{Webchild}, ConceptNet~\cite{ConceptNet}, and DBPedia~\cite{DBPedia}. In total, the dataset contains 13 relations: \textit {${\mathcal R}\in$ \{\textsc{Category}, \textsc{HasProperty}, \textsc{RelatedTo}, \textsc{AtLocation}, \textsc{IsA}, \textsc{HasA}, \textsc{CapableOf}, \textsc{UsedFor}, \textsc{Desires}, \textsc{PartOf}, \textsc{ReceivesAction}, \textsc{CreatedBy}, \textsc{Comparative}\}}. The dataset consists of 2190 images sampled from the ILSVRC~\cite{ILSVRC} and the MSCOCO~\cite{Lin2014} datasets. The accompanying KG consists of roughly 194500 facts, concerning 88606 entities. Roughly 82\% of the questions have a Key Visual Concept (KVC) as the answer, whereas 18\% have information from the Knowledge Base (KB) as the answer.

Answering questions in FVQA is to solve for
\begin{equation}
  \hat{y}=\argmax \limits_{e \in \mathcal{E}} \: p(y=e \: |\: q,I, \mathcal{G}),
  \label{eq:fvqa:a}
\end{equation}
i.e., finding the most probable entity as the answer given a question $q$ and image $I$, and given the graph $\mathcal{G}$. Our formulation of the missing edge reasoning task considers $\mathcal{G}_o$.

~\cite{FVQA} attempted FVQA as a parsing and querying problem, constructing 32 different templates of queries, and classifying each image-question pair as requiring one of these templates. Simple keyword matching techniques further prune the retrieved facts. 
Straight to the Facts (STTF) ~\cite{STTF} approached FVQA by directly learning to retrieve supporting facts, where each fact-entity was represented using lexical semantic representations.
~\cite{OOB} extended this approach in Out-of-the box (OOB), the current SOTA, by using local neighbourhood-based reasoning via a Graph Convolution Network (GCN) ~\cite{GCN} to answer each question. 
All of these current approaches rely on the ground-truth edge to be present in the KG. 

\section{Our approach - Seeing is Knowing}

The proposed architecture for FVQA is shown in Fig.~\ref{fig:Architecture}.  As shown, a given image $I$ and query $q$ are combined via CoAttention to form two entity query vectors, $f_{KVC}(q,I)$ and $f_{KB}(q,I)$.  The KG is then queried for the answer to the question, according to 
\begin{equation}
\small
  \hat{y}(q|I) = \begin{cases}
    \argmax \limits_{e\in\mathcal{E}} f_{KVC}(q,I)^T e & g_{KVC}(q)=1\\
    \argmax \limits_{e \in \mathcal{E}} f_{KB}(q,I)^T e & g_{KVC}(q)=0\end{cases}
    \label{eq:sikanswer}
\end{equation}
where the gating function $g_{KVC}(q)\in\left\{0,1\right\}$ is equal to 1 if the text of the question indicates that the answer is visible in the image, equal to 0 otherwise.  

The rest of this section addresses representations of the entities, image, and query, the
information fusion functions, the gating function, and
the loss function.

\subsection{KG representation}
KG Embeddings are learned by training a surrogate binary classifier to discriminate true edges from false. Since the KG only contains positive examples, we perform hard negative mining to generate negative examples using self-adversarial negative sampling \cite{RotatE}. A total of $n$ adversarial examples are generated for each positive example, and used to train discriminative embeddings using noise contrastive estimation~\cite{NCE}.
Thus the knowledge graph embedding  loss ${\mathcal L}_{KGE}$ includes the negative log probability that each observed edge is true ($\ln\sigma(\phi(f_i))$), and the expected log probability that the adversarial edges are false ($\ln\sigma(-\phi(f_j'))=\ln\left(1-\sigma(\phi(f_j'))\right)$):
\begin{equation}
\small
  {\mathcal L}_{KGE} = - \sum \limits_{i = 1}^ {|\mathcal{G}|}
  \left(\ln\sigma\big(\phi(f_i)\big)+
  E_i\left[\ln\sigma\big(-\phi(f_j')\big)\right]\right),
  \label{eq:LKGE}
\end{equation}
where expectation is with respect to the probability $p_i(f_j')$, tuned using
a temperature hyperparameter $\alpha$ as
\begin{equation}
  p_i(f_j') = \frac{\exp(\alpha\:\phi(f_j'))}{\sum\limits_{k=1}^n\exp(\alpha\:\phi(f_k'))}.
  \label{eq:Adversarial}
\end{equation}

Eq.~(\ref{eq:LKGE}) is used to train embeddings of the head ($h$) and tail ($t$), which are
applied to the FVQA task as described in the next several subsections.  
Eq.~(\ref{eq:LKGE}) also trains relation embeddings ($r$) and MLP weights for the ERMLP
scoring function ($w_{MLP}$); these quantities are not used for the downstream FVQA task.


\subsection{Image representation}
\cite{OOB} found that providing raw feature maps from a standard convolutional network like ResNet ~\cite{ResNet} or VGG~\cite{VGG} actually hurt performance on FVQA. We therefore represent each image using visual concepts found in the image.
	
\textbf{Objects:} We use Torchvision's Coco object-detector, a Faster RCNN detector ~\cite{FasterRCNN} with ResNet50 backbone ~\cite{ResNet}, and feature pyramid network ~\cite{FPN}, which detects 80 object classes. We also use a detector ~\cite{ZFTurbo} trained on an OpenImages 600-class detection task.  We include classes present in ImageNet 200 plus those in ~\cite{vanHengel} to maximize overlap with the dataset used with FVQA.
	
\textbf{Scenes:} We use a wideresnet ~\cite{WideResNet} detector trained on MIT365 places dataset ~\cite{MIT365} and consider the 205 classes that were used in constructing the FVQA KG.

Overall, we detect 540 visual concepts. 
	
	
Having detected visual concepts in each image, we represent the $i^{\textrm{th}}$ image as a collection of entities, $I_i=[e_i^1,\ldots,e_i^m]\in \Re^{N_e \times m}$, where $N_e$ is the embedding dimension, and $m$ is the number of visual concepts detected in the image. We detect a maximum of $m = 14$ visual concepts in each image. Our findings imply that for FVQA reasoning, an image is best represented as a collection of KG entity embedding vectors $e_i^j$, apparently because these KG embeddings encode the graph structure and background information necessary to be able to answer questions. We think this is an important finding, and we show its effect on the answer prediction accuracy. 

\begin{figure}[t]
  \centering
  \includegraphics[width=1\columnwidth]{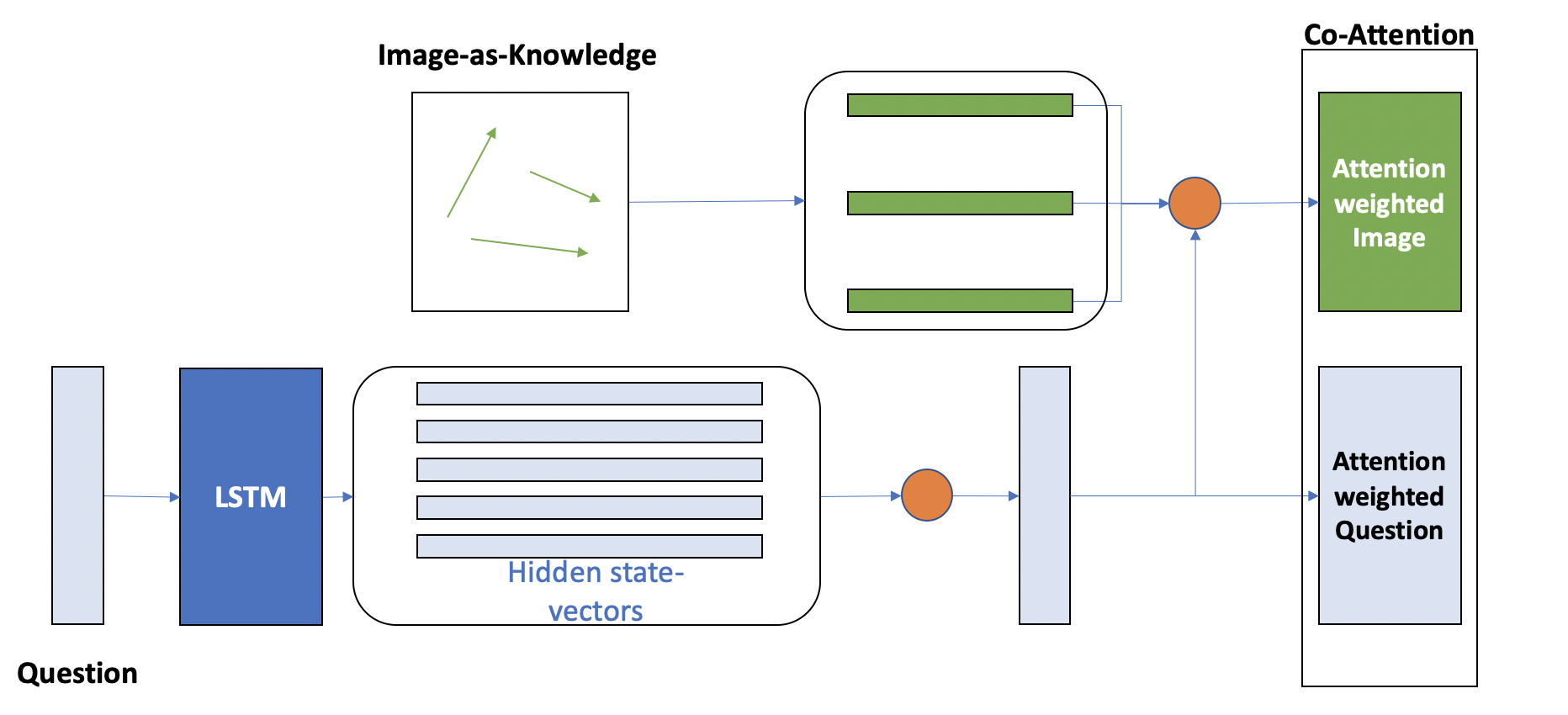}
  \caption{\small{Image  and query are fused using the CoAttention mechanism depicted here.  First, self-attention computes a weighted summary of the query (bottom orange circle).  Second, the query is used to compute an  attention-weighted summary of the concepts in the image (top orange circle).  The resulting image query is a vector drawn from the span of the entities present in the image.}}
  \label{fig:CoAttention}
\end{figure}
\subsubsection{Language representation}
For representing the question words, we use the last layer of contextual ALBERT ~\cite{ALBERT} embeddings without finetuning for FVQA training. After passing through an LSTM, we use the hidden state representation for each word $w_t$ for further processing as $q_i^t=h(w_t )$ as the question representation.

\begin{table*}[t]
  \centering
    \begin{tabular}{|P{5cm}|P{7cm}|P{2.5cm}| } 
      \hline
      \textbf{Parameters} & \textbf{Used for} & \textbf{Loss Function} \\
      \hline
      $h,r,t, w_{MLP}$  & KG embeddings & ${\mathcal L}_{KGE}$\\
      $w_{\alpha_q}$, $w_{\alpha_I}$, $w_{KVC}$, $w_{KB}$, $\theta_{LSTM}$  & Answer-retrieval based on image and question & ${\mathcal L}_{FVQA}$ \\         
      $w_{g}, \theta_{LSTM}$  & Choose answer source $\in\left\{f_{KVC},f_{KB}\right\}$ & Cross entropy\\                  
	\hline      
    \end{tabular}
    \caption{Summary of all trainable parameters.  Head
      ($h$) and tail ($t$) use identical embedding vectors.}
    \label{tab:table2}
\end{table*}
\begin{table*}[t]
  \centering
  \begin{tabular}{|P{3.5cm}|p{1.5cm}|P{1.5cm}|P{1.5cm}|P{1.5cm}|P{1.5cm}|P{1.5cm}|} 
    \hline
    \textbf{Method} & \textbf{Sampling} & \textbf{MR} & \textbf{MRR} & \textbf{Hits@1} & \textbf{Hits@3}& \textbf{Hits@10}\\
    \hline
    ERMLP & adversarial & $\textbf{11194}$  & $\textbf{0.156}$ & $\textbf{0.132}$ & $\textbf{0.152}$ & $\textbf{0.197}$ \\
    ERMLP & uniform & $14907$  & $0.122$ & $0.09$ & $0.128$ & $0.173$ \\
    \hline
    \multicolumn {7}{c}{\textbf{Incomplete KG results}} \\    
    \hline
    ERMLP (QA facts missing) & adversarial & $ 11071\pm479$  & $0.162\pm0.007$ & $0.136\pm0.003$ & $0.162\pm0.005$ & $0.204\pm0.007$ \\
    ERMLP (50\% KG) & adversarial & $12880$  & $ 0.144$ & $0.1$ & $0.13$ & $0.161$ \\
    \hline
  \end{tabular}
  \caption{KG Embedding accuracy for ERMLP (see Appendix for TransE and RotatE).}
  \label{tab:kge_results}
\end{table*}
    
\subsection{Fusion Functions $f_{KVC}$ and $f_{KB}$}
Image and query are fused using a CoAttention mechanism, as depicted in Fig.~\ref{fig:CoAttention}.  
First, we calculate attention weights for each word in the question, and subsequently a self-attention-weighted encoding for the question as:
\begin{equation}
A(q_i) = \sum \limits_{t =1}^ {|q_i|} \alpha_{q} ^t \: q_i ^t,~~~ \\
\alpha_{q} ^t = \frac{\exp(w_{\alpha_q} ^T q_i ^t)}{ \sum \limits_{t =1}^ {|q_i|} \exp(w_{\alpha_q} ^T q_i ^t)}
\label{eq: QuestionAttention}
\end{equation}
where $\alpha_q^t,w_{\alpha_q }$ are respectively the attention paid to word $w_t$, and the weight vector from which it is computed. 

Then, using $A(q_i)$ as a query, we compute an attention-weighted summary of the image:
\begin{equation}
\small
A(I_i) = \sum \limits_{j= 1}^m \alpha_{I} ^j \: e_i^j,~~~
\alpha_{I}^j = \frac{\exp\left(w_{\alpha_I} ^T \left[\begin{array}{c}A(q_i)\\e_i^j\end{array}\right]\right)}{\sum_{k=1}^m \exp\left(w_{\alpha_I}^T\left[\begin{array}{c}A(q_i)\\e_i^k\end{array}\right]\right)}
\label{eq:ImageAttention}
\end{equation}
where $\alpha_I^j,w_{\alpha_I}, e_i^j$ are respectively the attention paid to concept $j$ in the image, the weight vector from which it is computed, and the $j^{\textrm{th}}$ concept present in the image.
$A(I_i)$ is the attention-weighted representation of the image. 
Both $A(I_i)$ and $A(q_i)$ learn a mapping: $R^{N_e \times \: m} \rightarrow R^{N_e}$.


These attention-weighted image and question encodings compute joint image-question encodings via late fusion, as:
\begin{align}
  f_{KVC}(q_i, I_i) &= h\left(A(I_i),A(q_i);w_{KVC}\right)\\
  f_{KB}(q_i, I_i) &= h\left(A(I_i),A(q_i);w_{KB}\right)
  \label{eq:fkb}
\end{align}
where $h(\cdot)$ is a two-layer fully-connected network with ReLU activation functions.
Using $f_{KVC}(q_i, I_i) = A(I_i)$, i.e., the attention weighted image encodings to directly retrieve the answer significantly reduces accuracy, suggesting the need for successive fully connected layers to add capacity.

\subsection{Gating function $g_{KVC}$}  The fusion functions $f_{KVC}$ and $f_{KB}$ are trained to retrieve two different entities from the KG, either of which might be the answer to the question.  The gating function, $g_{KVC}$, selects one of these two.  The gating function is a sigmoid fully-connected layer applied to the final output state of an LSTM, whose input is the query, and which is trained using binary cross entropy so that $g_{KVC}=1$ if the correct answer is a key visual concept in the image.  

\subsection{Loss function}
A summary of all parameters is provided in Table~\ref{tab:table2}.
Entity embeddings are learned in order to minimize the loss function in Eq. (\ref{eq:LKGE}),
then all parameters are jointly trained in order to minimize the cosine distance between the ground truth entity, $y_i$, and the network output:
\begin{equation}
  {\mathcal L}_{FVQA}=\frac{1}{n}\sum_{i=1}^n\left(1-y_i^T\hat{y}(q_i|I_i)\right)
 \label{eq:LFVQA}
\end{equation}
where $\hat{y}(q_i|I_i)$ is as given in Eq. (\ref{eq:sikanswer}).

\section{Text-augmented composite score}
\subsection{Text-augmented score metric}
\begin{algorithm}
\SetAlgoLined
\For {$q_i, I_i \in \mathcal D$}
{Obtain token set $ \mathcal T = \texttt{TOKENIZE}(q_i,I_i)$\;
	\For {$f_k \in \mathcal{G}$}
	{
		\For {$w_t \in \texttt{TOKENIZE}(f_k)$}			
		{
			$S(w_t) = \underset{c \in \mathcal T} \max \frac{g(w_t)^T g(c)}{|g(w_t)| |g(c)|}$\;
		}
		$\mathcal M = \{w_j: \underset{j \in |f_k|} {\texttt{argsort}} \ S(w_j) \} $;\\
		$\eta(f_k) = \sum_{j=1}^{0.8*|\mathcal M|} S(w_j) $\;
    }
    $\mathcal F_{100} = \{f_k: \underset{k \in |\mathcal G|}  {\texttt{argsort}} \ \eta(f_k) \}_{k = 1}^{100} $\;
} 

\caption{Pruning entity search space to set of most relevant facts $\mathcal F_{100}$}
\label{alg:Search}
\end{algorithm}
To enhance SiK's answer retrieval, we define additionally a text-augmented metric to leverage lexical similarity between the facts and the question in the algorithm as described in \ref{alg:Search}. Firstly, a pruning strategy as described in \cite{OOB} is implemented. Using GLoVE 100-D embeddings, a similarity score is computed for each KG fact, by first computing cosine similarity for each word in the fact with every word in the question and detected visual entities. The top 80\% scoring words from each fact are retained, and these similarity scores are averaged to assign a score to each fact for a given question. The top 100 scoring facts are retrieved for each question, denoted as ${\mathcal F}_{100}$. Answer retrieval then takes place from entities within this reduced set of facts ${\mathcal F}_{100}$. We define the score function, a convex combination of three complementary metrics, as follows:
\begin{align*}
 \hat{y}(q|I) &= \argmax \limits_{e\in {\mathcal F}_{100}}\left\{\lambda_1 K(q, I)^Te + \lambda_{2} {J(q,e)}\right.    \\
& \left. +\lambda_3 G(q, e)\right\}~~~~\mbox{s.t.}~\sum_{k=1}^{3} \lambda_k = 1
\label{eq:composite}
\end{align*}
Here, $K(q,I)^T e$ is the KG similarity score of an entity with the query produced by the SiK network. $J(q,e) = \max_{f \in {\mathcal F}_{100},e\in f} j(q, f)$, where $j(q,f)$ is the Jaccard (word-token) similarity between the question and one of the retrieved facts $f$.  An entity may be a part of more than one fact in ${\mathcal F}_{100}$; $e\in f$ means that $e$ is either the head or tail of $f$.  $G(q,e) = \max_{f \in {\mathcal F}_{100},e\in f}g(q, f)$ where $g(q, f)$ is the similarity between its fact and a question based on averaged GloVe embeddings.  
\begin{table*}[t]
  \centering
  \begin{tabular}{|P{8cm}|P{3cm}|P{3cm}|} 
    \hline
    \textbf{Technique} & \textbf{Hits@1} & \textbf{Hits@3}\\
    \hline
    HieCoAtt & $33.7 \pm 1.18$ & $50.00 \pm 0.78$ \\
    FVQA top-1 qq-mapping &$52.56 \pm 1.03$ & $59.72 \pm 0.82$ \\
    STTF & $62.2 \pm NR$ & $75.6 \pm NR$ \\
    OOB top-1 rel & $\mathbf{65.8 \pm NR}$ & $\mathbf{77.32 \pm NR}$ \\
    OOB top-3 rel & $69.35 \pm NR$ & $80.25 \pm NR$\\
    \hline
    \multicolumn {3}{c}{\textbf{Seeing is Knowing}} \\
    \hline
    AvgEmbed, IaK & $27.77 \pm 0.72$ & $32.43 \pm 0.94$\\
    ERMLP – Adv, Multihot & $51.51 \pm 1.14$ & $64.89 \pm 1.37$\\
    ERMLP, IaK & $53.16 \pm 0.79$ & $63.9 \pm .63$\\
    ERMLP -Adv, IaK  & $\mathbf{54.38 \pm 0.94}$ & $\mathbf{65.76\pm 0.5}$\\
    \hline
	\multicolumn {3}{c}{\textbf{Seeing is Knowing - text augmented}} \\
    \hline    
	ERMLP - Adv, IaK ($\lambda_1 = 0.4$, $\lambda_2 = 0.3$,$\lambda_3=0.3$)& $\mathbf{60.82\pm0.94}$ &$\mathbf{77.14\pm0.50}$ \\
	ERMLP - Adv, IaK ($\lambda_1 = 0$, $\lambda_2 = 0.5$,$\lambda_3=0.5$)& $39.76 \pm1.08 $  & $60.84 \pm 0.67 $ \\
	\hline
	\end{tabular}
\newline	
 \begin{tabular}{|P{5cm}|P{2.2cm}|P{2.2cm}|P{2.2cm}|P{2.2cm}|}
    \multicolumn {5}{c}{\textbf{Incomplete KG}} \\
    \hline
    \textbf{Method}& \multicolumn {2}{c|}{\textbf{Only QA facts missing}} &\multicolumn {2}{c|}{\textbf{50\% KG missing}} \\
    \hline
	AvgEmbed, IaK& $27.77 \pm 0.72$ & $32.43 \pm 0.94$ &$27.77 \pm 0.72$ & $32.43 \pm 0.94$\\\
    OOB top-3 rel &$28.58 \pm 0.01$ & $43.56 \pm 0.01$&$27.53 \pm 0.01$ & $41.56 \pm 0.01$  \\
    ERMLP -Adv, IaK & $ 53.45 \pm 0.77$  & $65.1 \pm 1.41$ &$52.29 \pm 0.86$ & $62.74 \pm 0.69$ \\
    ERMLP - Adv, IaK ($\lambda_1=0.34$, $\lambda_2=0.33$,$\lambda_3=0.33$) &$\mathbf{55.13\pm0.97}$ &$\mathbf{73.04\pm0.54}$ &$\mathbf{55.14\pm1.1}$&$\mathbf{72.95\pm0.38}$ \\
    ERMLP - Adv, IaK ($\lambda_1=0$, $\lambda_2=0.5$,$\lambda_3=0.5$) &$26.8\pm0.90$&$49.24\pm0.62$&$26.78\pm0.856$&$49.20\pm0.63$ \\
    \hline
    
  \end{tabular}
  \caption{FVQA accuracy (IaK - Image as Knowledge, Adv - self-adversarial negative sampling, NR - Not reported)}
  \label{tab:fvqa_results}  
\end{table*}
\section{Experimental Setup}
\subsection{KG Embeddings training}
For training of each KG embedding algorithm, we split all the facts in the KG as 80\% train, 20\% test. Entity and relation embeddings have a dimension of $N_e= 300$. Batch size is 1000. Each network is trained for 25000 steps using the Adam optimizer with learning rate initallized as 0.01 shrunk every 10000 steps by 0.1. Learning rates for all experiments were decided by choosing from the set \{0.1, 0.01. 0.001, 0.0001\}. We set $\alpha = 1$ for all experiments. We use $n=16$ negative samples for each positive sample.  ERMLP uses a three layer neural network with $3N_e$ (taking in head, rel, and tail embeddings), $2N_e=600$ and $N_e=300$ units respectively, and the output contains one sigmoid unit (the score $\phi (h, r, t)$). Hidden activations are ReLU. No dropout was used. 

KGE accuracy is measured using standard metrics: Hits @1, Hits @3, Hits @10 determine how often each correct tail/head gets ranked in the top 1, 3, or 10 ranked facts for each ground-truth $(h, r)$/$(r,t)$ pair. Mean Rank measures the mean rank of each true fact in the dataset, Mean Reciprocal Rank = $\frac{1}{|{\mathcal D}|}\Sigma_i\frac{1}{R_i}$. While we report these metrics, we do note that these metrics are not best suited for common-sense KGs for reasons mentioned before. However we leave a more refined analysis of these methods for future work. 

\subsection{FVQA training}
We report Hits @1 and Hits @3 for each algorithm. All numbers are based on averaging results across five train - test splits provided with the dataset. The number of questions varies slightly, but roughly fall half into both training and testing sets for each split. Stochastic gradient descent with a batch size of 64 trains $f_{KVC}$ and $f_{KB}$ for 250 steps with a learning rate of 0.01, reduced by 0.1 every 100 epochs, and a weight decay of 1e-3. Fully-connected layers use a dropout probability of 0.3. The gating function $g_{KVC}$ is trained for 20 steps with step-size of 0.1. 

GPUs provided by Google Colab are used to train all models. Our heaviest KG embedding technique (ERMLP) takes roughly 3 hours to train, while one train split for $f_{KVC}$ takes roughly 30 minutes. Subgraph construction for each queston takes about 1s, and 250 epochs to train our implementation of OOB took roughly 3 hours.

\subsection{Baselines}
Results in the standard QA task for the models FVQA, STTF, and OOB are taken from the respective papers since the codes for these systems have not been made available. HieCoAtt denotes using a Hierarchical CoAttention network  \cite{HieCoAtt} as implemented in \cite{FVQA}. AvgEmbed denotes a method which compares STTF's and OOB's fact representation methodology with our architecture to correctly perform missing-edge reasoning. We use averaged 300-dimensional GLoVE embeddings ~\cite{GloVe} to represent each entity. We also report results from our OOB implementation in the incomplete KG setting. 


\section{Results and Discussion}
The gating function achieves an accuracy of 96\%, same as reported in ~\cite{STTF}. We discuss below the performance of other modules of our architecture. 

\subsection{${\mathcal F}_{100}$ fact recall and impact on FVQA}
We only observe 63\% fact recall within ${\mathcal F}_{100}$ across the entire dataset as opposed to 84.8\% as reported by \cite{OOB}. This further goes down to 53\% if filtering for top-3 relationships. This difference is significant, and FVQA accuracy reported by OOB at such levels of fact recall are lower. One factor for lower recall is noisy visual detections, whose keywords impact fact retrieval. Therefore, with access to similar visual detections as the other methods, SiK could potentially achieve even higher accuracy. Furthermore, owing to the lower fact retrieval recall, our OOB implementation could not replicate the reported performance in the standard QA task. This also highlights the crucial reliance of OOB on correct fact retrieval and further substantiates our observation of its low performance when the KG is incomplete. 
\begin{figure*}[ht]
    \begin{tabular}{p{5cm} p{5cm} p{5cm}}
    \includegraphics[width=.62\columnwidth, height = 0.5\columnwidth.]{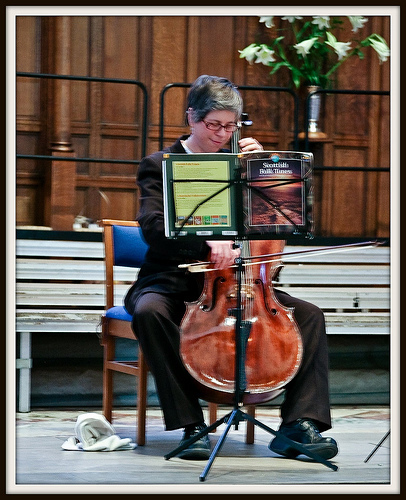} &
    \includegraphics[width=.62\columnwidth, height = 0.5\columnwidth.]{Success3} &
    \includegraphics[width=.62\columnwidth, height = 0.5\columnwidth.]{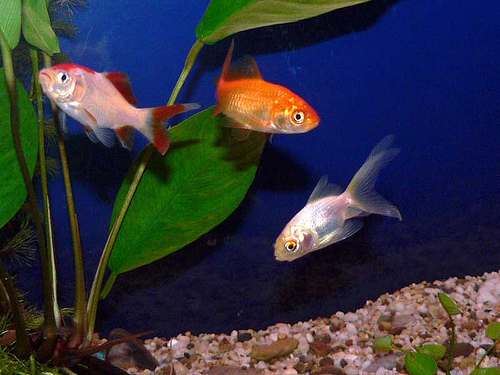} \\
      \textbf{Question 1:} What is the difference between the instrument \& the violin? \newline
      \textbf{SPO triple:} : \{Cello, \textsc{HasProperty}, like a violin but larger\}  \newline
	  \textbf{Answer Source:} KG \newline
	  \textbf{Answer:} like a violin but larger \newline
	  \textbf{Answer predicted: \textcolor{OliveGreen}{like a violin but larger}} &
  
      \textbf{Question 2:} Which object in this image belongs to the category Herbivorous animals? \newline
      \textbf{SPO triple:} \{Giraffe, \textsc{Category}, Herbivorous animals\} \newline
	  \textbf{Answer Source:} Image\newline
	  \textbf{Answer:} Giraffe \newline
	  \textbf{Answer predicted: \textcolor{OliveGreen}{Giraffe}} &
	  
	  \textbf{Question 3:} What popular pet is in this image? \newline
      \textbf{SPO triple:} \{Goldfish, \textsc{IsA}, popular pet\} \newline
	  \textbf{Answer Source:} Image \newline
	  \textbf{Answer:} Goldfish \newline
	  \textbf{Answer predicted: \textcolor{Red}{Fish}} \\
	  
	  \includegraphics[width=.62\columnwidth, height = 0.5\columnwidth.]{Failure2} &
    \includegraphics[width=.62\columnwidth, height = 0.5\columnwidth.]{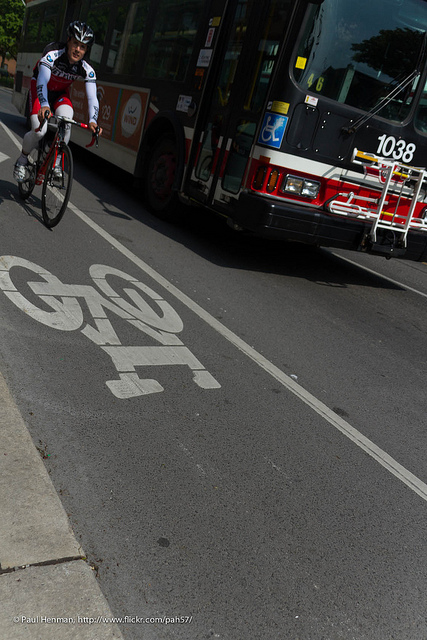} &
    \includegraphics[width=.62\columnwidth, height = 0.5\columnwidth.]{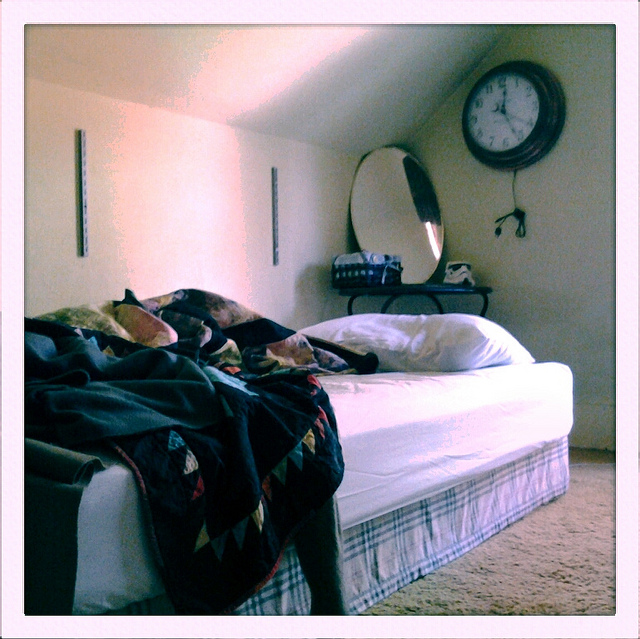} \\

      \textbf{Question 4:} What object in this image is commonly eaten for lunch?\newline
      \textbf{SPO triple:}  \{Sandwich, \textsc{IsA}, meal commonly eaten for lunch\} \newline
	  \textbf{Answer Source:} Image \newline
	  \textbf{Answer:}Sandwich \newline
	  \textbf{Answer predicted: \textcolor{Red}{Bread}} &
    	  
	  \textbf{Question 5:}Where can you find the large object in the back of this image?   \newline
      \textbf{SPO triple:}\{Bus, \textsc{AtLocation}, Bus stop\}  \newline
	  \textbf{Answer Source:} KG \newline
	  \textbf{Answer:} Bus stop  \newline
	  \textbf{Answer predicted: \textcolor{Red}{Wait place}} &
	  
      \textbf{Question 6:} What can we find in the place shown in this image? \newline
      \textbf{SPO triple:} \{Furniture, \textsc{AtLocation}, Bedroom\}  \newline
	  \textbf{Answer Source:} KG\newline
	  \textbf{Answer:}Furniture  \newline
	  \textbf{Answer predicted: \textcolor{Red}{Your House}} \\

    \end{tabular}    

    \caption{Success and Failure cases of Seeing is Knowing}
    \label{fig: Examples}
\end{figure*}
\subsection{Incomplete KGs}
The robustness of SiK is demonstrated in the incomplete KG setting. Here we discuss two of them - one where only the QA-related facts are missing and another where 50\% of the KG is missing (more settings in Appendix). In both cases, we see standalone SiK and text-augmented SiK only lose some of their accuracy. Our implementation of OOB, and AvgEmbed both underperform SiK by over 25\%. Setting $\lambda_1 = 0$ in the score also leads to similar performance, highlighting the importance of KG features for FVQA. For a more detailed discussion of the missing edges experiments, the interested reader is directed to the Appendix.

\subsection{Composite metric performance}
When the required edge is present in the graph, standalone SiK underperforms OOB by 11\%. This is enhanced using the composite score, which sees a 6\% point improvement. Quite remarkably, text-augmented SiK works best when roughly equal coefficients are used for all three similarities. However, the strongest similarity metric is still that obtained from SiK (see Table \ref{tab:fvqa_results} and Appendix). 

\subsection{Image as Knowledge}

To understand how useful the Image-as-Knoweldge representation is compared to other variants, we compare it to the multihot variant proposed by \cite{OOB} along with our best performing KG embedding technique.
`Image-as-knowledge' provides a 3-point performance improvement (Table~\ref{tab:fvqa_results}), apparently
because the retrieval of an entity happens in the entity space spanned by the vectors of the IaK representation.

\subsection{Success and Failure cases}
To investigate the only learning module in the score, we look at the failure modes for the standalone SiK architecture. Since the gating function performs with almost perfect accuracy, our main source of error is incorrect entity prediction. Since a prediction is considered correct only if there's an exact match in the entities, a lot of inaccuracies stem from cases where the network predicts a semantically valid but different entity. Entities sharing parent-child relationships (Question 3, Fig. \ref{fig:Example}) or similar attributes through multi-hop relationships (Questions 4-6) are commonly mistaken by the SiK network. We discuss more failure cases in the appendix. 
\section{Conclusion}
An architecture is proposed that solves the FVQA task even when the required edge is missing from the knowledge graph. In the process, we present the first approach to use KG embeddings for FVQA. A composite answer retrieval method augments its accuracy by incorporating complementary lexical features and KG semantic features. Serendipitous benefits of the standalone approach include: (1) improved computational complexity, and (2) an improved representation of each image, as the span of KG embeddings of the visible entities. 

\section{Ethical impact}
\begin{figure}[ht]
        \centering
      \includegraphics[width=0.6\columnwidth,height=0.7\columnwidth]{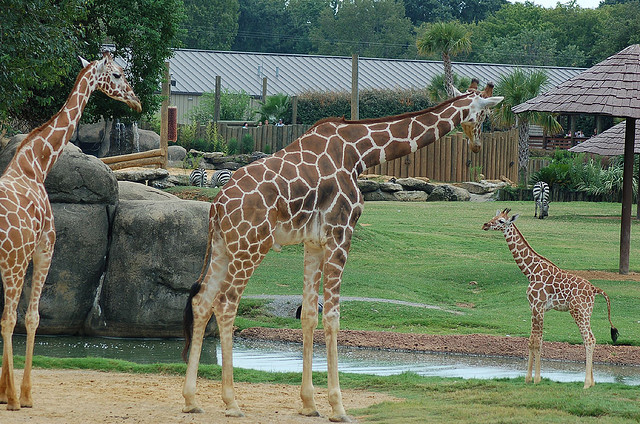}
     \captionof{figure}{%
     \small{
  \textbf{Question:} Where is this place? \\
  \textbf{SPO triple:} : \{Life, \textsc{AtLocation}, Zoo\} \\
  \textbf{Answer Source:} Knowledge Base \\
  \textbf{Answer:} Zoo \\
  \textbf{Answer predicted: \textcolor{Red}{in Zimbabwe}} \\
     }}
     \label{fig:Ethics}
\end{figure}

We now examine the ethical implications of our work. A prominent issue could be that of different biases known to exist in our data sources. ~\cite{ImageBias} show population / representation bias existing in OpenImages and ImageNet. ~\cite{SocialBias} showed web-scale common-sense KGs can be tough to curate and can allow biases to creep in. ~\cite{DBPediaBias} note how the density of world locations generating DBPedia data (extracted from Wikipedia) is at odds with world population density. Fig. ~\ref{fig:Ethics} shows how this manifests in our system. The answer for a place relevant to giraffes in wildlife, comes up as Zimbabwe, even though the only edges present in the KG concerning Zimbabwe were \{Person, \textsc{AtLocation}, in Zimbabwe\}, \{Dog, \textsc{AtLocation}, in Zimbabwe\}, \{Tree, \textsc{AtLocation}, in Zimbabwe\}, \{Elephant, \textsc{AtLocation}, in Zimbabwe\}. Such an error is quite informative and humbling – this means that different modalities (image, language, graph) working in tandem could still amplify their individual biases. For deploying such a system, we recommend debiasing of parameters learned in our architecture. While ConceptNet has been active in debiasing their representations, debiasing KG embeddings has not received as much attention, and it could pose subtle problems given that most entities would have low node-degree. A multimodal approach to debias representations could be an interesting research direction.

\bibliographystyle{acl_natbib}
\bibliography{mybib}

\section{Appendix}
\pdfinfo{
/Title (Seeing is Knowing! Fact-based VQA using Knowledge Graph Embeddings )
/Author (Kiran Ramnath, Mark Hasegawa-Johnson)
/TemplateVersion (2021.1)
} 


\author{First Author \\
  Affiliation / Address line 1 \\
  Affiliation / Address line 2 \\
  Affiliation / Address line 3 \\
  \texttt{email@domain} \\\And
  Second Author \\
  Affiliation / Address line 1 \\
  Affiliation / Address line 2 \\
  Affiliation / Address line 3 \\
  \texttt{email@domain} \\}

\date{}

\pdfinfo{
/Title (Seeing is Knowing! Fact-based VQA using Knowledge Graph Embeddings )
/Author (Kiran Ramnath, Mark Hasegawa-Johnson)
/TemplateVersion (2021.1)
} 

\setcounter{secnumdepth}{0} 

%



\title{Seeing is Knowing - Supplementary material}

\maketitle

\begin{table*}[ht]
  \centering
	\begin{tabular}{|P{1.7cm}|P{2.3cm}|P{1.8cm}|P{1.8cm}|P{1.8cm}|P{1.8cm}|P{1.8cm}|}
    \hline
    \textbf{Method} & \textbf{KG Occlusion} & \textbf{MR} & \textbf{MRR} & \textbf{Hits@1} & \textbf{Hits@3}& \textbf{Hits@10}\\
    \hline
    ERMLP & Only QA facts & $ 11071$  & $0.162$ & $0.136$ & $0.162$ & $0.204$ \\
    ERMLP & 10\% of KG & $9690$  & $0.175$ & $0.148$ & $0.1779$ & $0.22$ \\
    ERMLP & 20\% of KG & $10733$  & $0.1613$ & $0.133$ & $0.162$ & $0.213$ \\
    ERMLP & 30\% of KG & $10437$  & $0.154$ & $0.127$ & $0.156$ & $0.199$ \\
    ERMLP & 40\% of KG & $11467$  & $0.139$ & $0.116$ & $0.138$ & $0.179$ \\
    ERMLP & 50\% of KG & $12880$  & $ 0.144$ & $0.1$ & $0.13$ & $0.161$ \\
    \hline
  \end{tabular}
  \caption{KG Embedding performance of ERMLP when varying levels of the KG is occluded.}
  \label{tab:kge_missing}
\end{table*}

\begin{table*}[ht]
  \centering
 \begin{tabular}{|P{4cm}|P{4cm}|P{3cm}|P{3cm}|}
    \hline
	\textbf{ERMLP} & \textbf{KG Occlusion} & \textbf{Hits @1} & \textbf{Hits @3}\\
    \hline

	ERMLP & QA facts & $ 53.45 \pm 0.77$  & $65.1 \pm 1.41$  \\
	ERMLP &10\% of KG & $ 53.19 \pm 0.47 $  & $65.67 \pm 0.31 $  \\
   ERMLP & 20\% of KG & $ 53.62 \pm 1.01$  & $ 64.72 \pm 0.98$  \\
    ERMLP &30\% of KG & $ 53.15 \pm 1.62$  & $ 64.55 \pm 1.23$  \\
    ERMLP &40\% of KG & $ 53.14 \pm0.47$  & $64.77 \pm 0.55 $  \\
	ERMLP &50\% of KG & $ 52.29 \pm 0.86 $  & $62.74 \pm 0.69 $  \\
    \hline
  \end{tabular}
  \caption{FVQA accuracy of Seeing is Knowing using ERMLP when varying levels of the KG is occluded.}
  \label{tab:fvqa_missing}
\end{table*}

\section{A. Experiments with incomplete KG}
Table ~\ref{tab:kge_missing} shows the performance of KG Embeddings using standard metrics at different levels of KG incompleteness. As expected, there is a small but monotonous decline in the performance with more and more occlusion. A known shortcoming of these metrics is that when the KG has $(subject, predicate)$ pairs that can lead to more than one $object$ entity, they will consider a different but factually correct entity as incorrect. 

\if
This corresponds to the case of setting $\lambda_1 = 1, \lambda_2 = 0, \lambda_3 = 0$
\fi

Quite remarkably however, there is only a very slight decrease in SiK's performance when using any of these KG embeddings for the downstream FVQA task (see Table \ref{tab:fvqa_missing}). Only at 50\% KG occlusion, we see a statistically significant yet still small drop in performance. This highlights the robustness of using KG embeddings for down-stream semantic tasks. 

\begin{figure*}
    \begin{tabular}{p{5cm} p{5cm} p{5cm}}

    \includegraphics[width=.62\columnwidth, height = 0.5\columnwidth.]{supplements/Giraffe} &
    \includegraphics[width=.62\columnwidth, height = 0.5\columnwidth.]{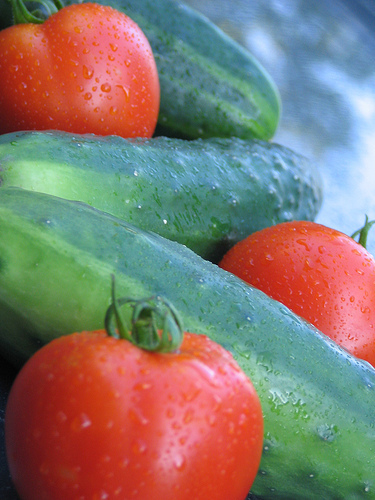} &
    \includegraphics[width=.62\columnwidth, height = 0.5\columnwidth.]{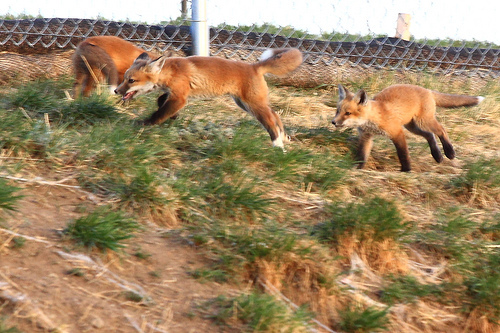} \\
      \textbf{Question 1:} What is the animal famous for?   \newline
      \textbf{SPO triple:} \{Giraffe, HasProperty, Long necked\} \newline
	  \textbf{Answer Source:} KG  \newline
	  \textbf{Answer:}long necked \newline
	  \textbf{Answer predicted: \textcolor{Red}{work for day without water}} &

      \textbf{Question 2:} Which object in this image is used in spaghetti sauce?  \newline
      \textbf{SPO triple:} \{Sphagetti sauce, RelatedTo, Tomato\}  \newline
	  \textbf{Answer Source:} Image \newline
	  \textbf{Answer:} Tomato \newline
	  \textbf{Answer predicted: \textcolor{Red}{Bell Pepper}} &	    

      \textbf{Question 3:} What is the family of the animal in this image?  \newline
      \textbf{SPO triple:}\{Fox, Category, Canidae\}  \newline
	  \textbf{Answer Source:} KG \newline
	  \textbf{Answer:}Canidae  \newline
	  \textbf{Answer predicted: \textcolor{Red}{Burundian Culture}} \\
	  
    \includegraphics[width=.62\columnwidth, height = 0.5\columnwidth.]{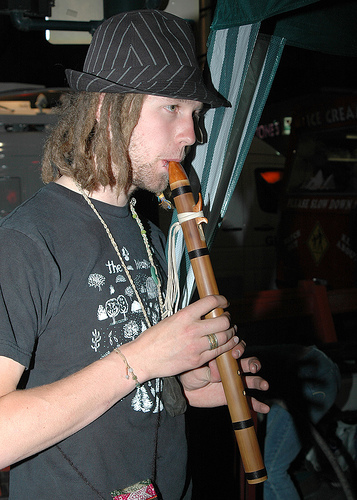} &
    \includegraphics[width=.62\columnwidth, height = 0.5\columnwidth.]{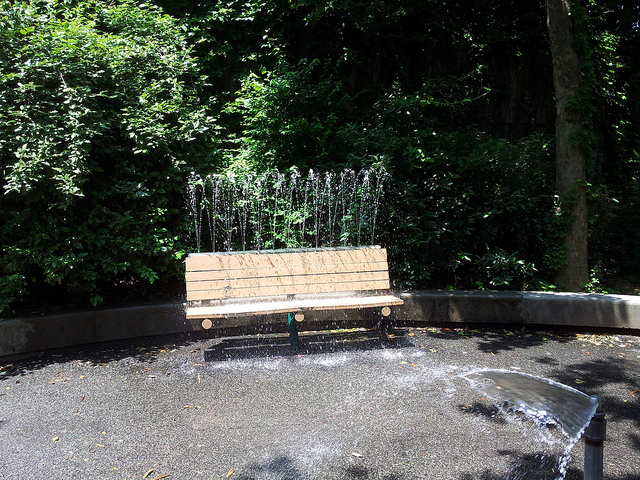} &
    \includegraphics[width=.62\columnwidth, height = 0.5\columnwidth.]{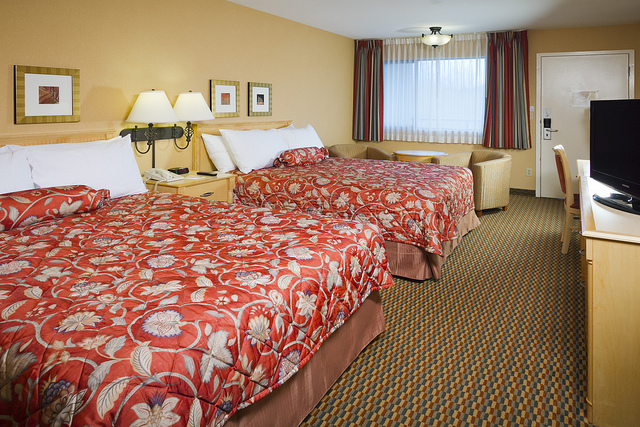} \\
      \textbf{Question 4:} What is the wooden thing used for?   \newline
      \textbf{SPO triple:} \{Flute, UsedFor, making music\}\newline
	  \textbf{Answer Source:} KG  \newline
	  \textbf{Answer:}making music \newline
	  \textbf{Answer predicted: \textcolor{Red}{Practice}} &
	      
	  \textbf{Question 5:} What is the material of the chair?\newline
      \textbf{SPO triple:} \{Chair, RelatedTo, Wooden\} \newline
	  \textbf{Answer Source:} KG \newline
	  \textbf{Answer:}\newline
	  \textbf{Answer predicted: \textcolor{Red}{come in many style}} &	  		
	  	  
	  \textbf{Question 6:} What is the object in the left of the image used for?  \newline
      \textbf{SPO triple:}\{Bed, UsedFor, Laying on\} \newline
	  \textbf{Answer Source:} KG  \newline
	  \textbf{Answer:} Laying on  \newline
	  \textbf{Answer predicted: \textcolor{Red}{Loose (sic) thing in}} \\

    \includegraphics[width=.62\columnwidth, height = 0.5\columnwidth.]{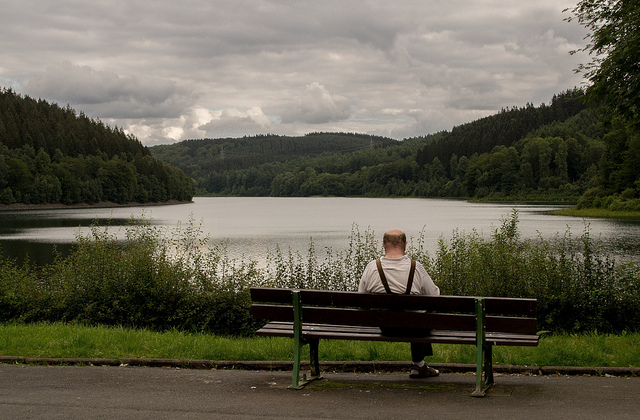} &
    \includegraphics[width=.62\columnwidth, height = 0.5\columnwidth.]{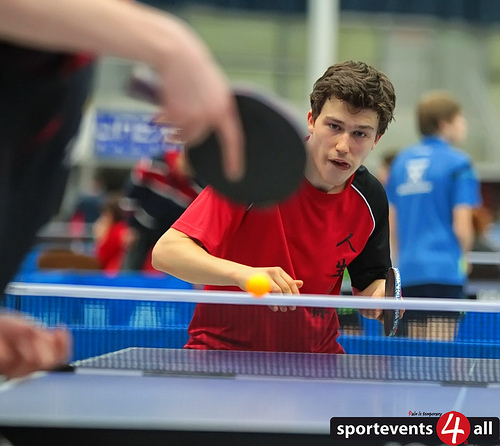} &
    \includegraphics[width=.62\columnwidth, height = 0.5\columnwidth.]{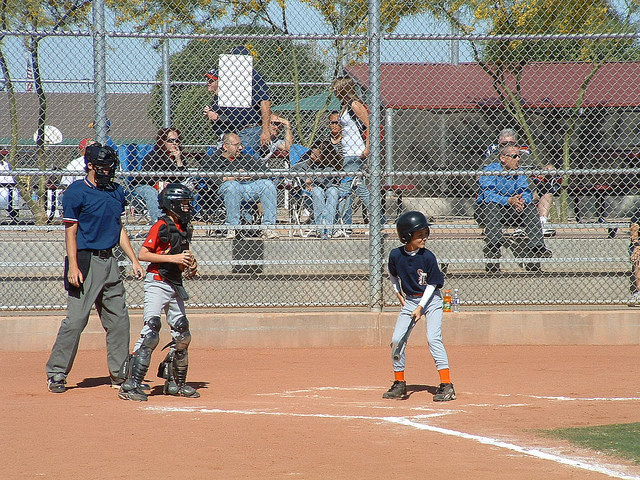} \\
   
      \textbf{Question 7:} What can we do in this place shown in this image? \newline
      \textbf{SPO triple:}\{Lake, UsedFor, row a boat\} \newline
	  \textbf{Answer Source:} KG   \newline
	  \textbf{Answer:} row a boat  \newline
	  \textbf{Answer predicted: \textcolor{Red}{build a treehouse}} &
	      
      \textbf{Question 8:} This game is most popular in which country? \newline
      \textbf{SPO triple:}\{Table Tennis, HasProperty, Popular in China\} \newline
	  \textbf{Answer Source:} KG \newline
	  \textbf{Answer:}Table Tennis  \newline
	  \textbf{Answer predicted: \textcolor{Red}{Game Person Play}} &      
	  	   	  	  
	  \textbf{Question 9:} What the baseball bat looks like? \newline
      \textbf{SPO triple:} \{Baseball, IsA, Long round tapered object\} \newline
	  \textbf{Answer Source:} KG  \newline
	  \textbf{Answer:} long round tapered object \newline
	  \textbf{Answer predicted: \textcolor{Red}{aggressive animal}} \\		
\end{tabular}
\caption{Performance of standalone Seeing is Knowing network without the composite score-based retrieval}	  	  	 	   
\label{fig:examples}
\end{figure*}	    

\section{B. Characterizing inaccuracies of Seeing is Knowing Network: Qualitative discussion}
In order to analyze the types of errors SiK is prone to make, we now discuss some more failure cases. We observe that there are three broad categories of errors -  \\

\begin{enumerate}
\itemsep-1em
\item Semantically correct answer, but answer is a different entity \\
\item Incorrect answer, explainable relevance to question and image \\
\item Entirely incorrect answers \\
\end{enumerate}

An empirical study of these errors shows a significant fraction of them fall in the first two categories. Errors in the first category can arise when the model has implicitly pinned down the correct key visual entity and relationship being asked about, but then chose a different entity (see Question 4-6 Fig.\ref{fig:examples}). A bed is an object one can lose things in, chairs come in many styles, and a flute can be used for practice. These errors can also arise if the model has chosen to place visual emphasis on an entity that is different from ground truth, but is equally valid given a question (see Question 7 Fig.\ref{fig:examples}). Here, the model has retrieved the answer `build a treehouse' that appears in a fact with `tree', but the ground truth of the question asks about the entity `lake'. 

Errors of the second category can arise in mainly two cases - incorrect language and / or visual grounding. In cases where the visual grounding is off, it could be due to false positives in detections (see Questions 1-3 Fig.\ref{fig:examples}).  Giraffes, tomatoes, and foxes are mistaken to be camels, bell peppers, and lion respectively; entities relevant to the mistaken entities are then chosen as answers. 

In cases where the language grounding is off, one case could be that the model has predicted the correct symbolism for a question (i.e. correct entity and relationship), but the retrieved entity comes from a fact that is not grounded in the language of the question. In Question 8 Fig. ~\ref{fig:examples}, `game person play', a property about the game table tennis is chosen, but it is not grounded in the question-text which asks about the country where it is popular. A fix for this would be to ensure that entities' representations contain both lexical and graph-based features. One approach could be to use contextual word embeddings to encode entities before doing graph learning. Another approach involving RotatE, could be to fix real part of the entities as lexical representations during training, so that the imaginary parts could be enforced to encode graph-based features for fixed lexical semantic features. 

Lastly, entirely incorrect answers, such as Question 9 Fig.\ref{fig:examples} can occur in case of sparse node connections (only one edge is present for `aggressive animal'). SiK is indeed prone to making such unexplainable errors, highlighting the scope for improvement to be gained from incorporating better lexical semantic features as used in our composite score function. 

\section{C. Attention maps}
\begin{figure*}[ht]
    \begin{tabular}{p{5cm} p{5cm} p{5cm}}

    \includegraphics[width=.62\columnwidth, height = 0.5\columnwidth.]{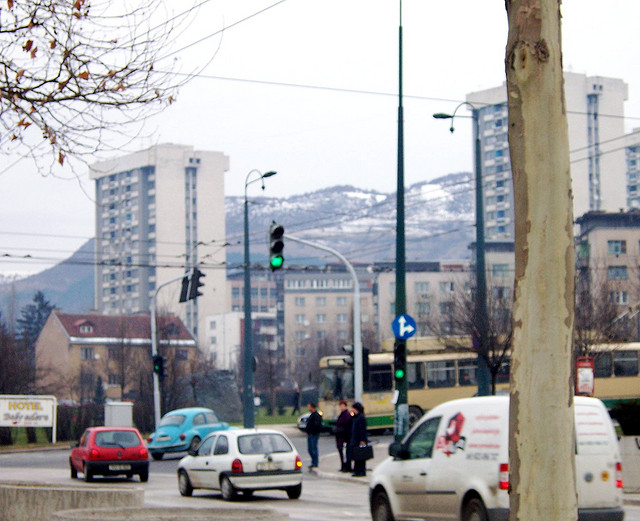} &
    \includegraphics[width=.62\columnwidth, height = 0.5\columnwidth.]{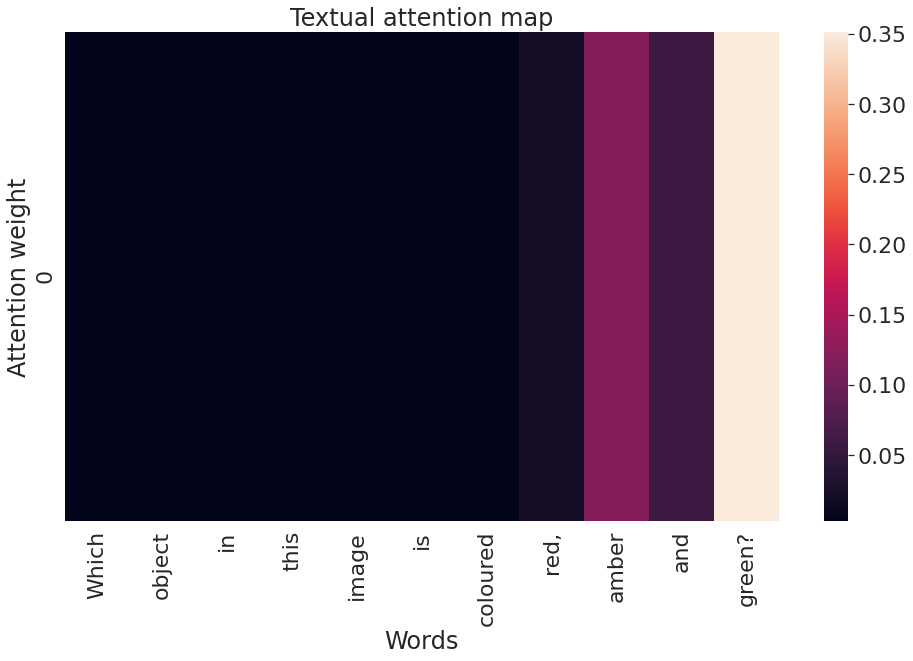} &
    \includegraphics[width=.62\columnwidth, height = 0.5\columnwidth.]{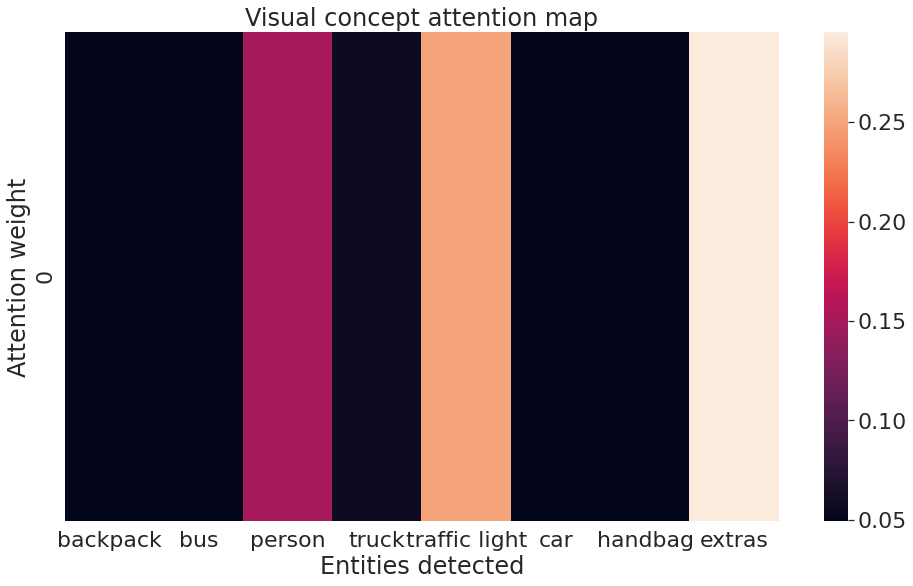} \\
      \textbf{Question 1:} Which object in this image is colored red, amber and green?  \newline
      \textbf{SPO triple:} \{Traffic Light, HasProperty, colour red amber and green\} \newline
	  \textbf{Answer Source:} Image  \newline
	  \textbf{Answer:} Traffic light\newline
	  \textbf{Answer predicted: \textcolor{Green}{Traffic light}} 	 & & \\

    \includegraphics[width=.62\columnwidth, height = 0.5\columnwidth.]{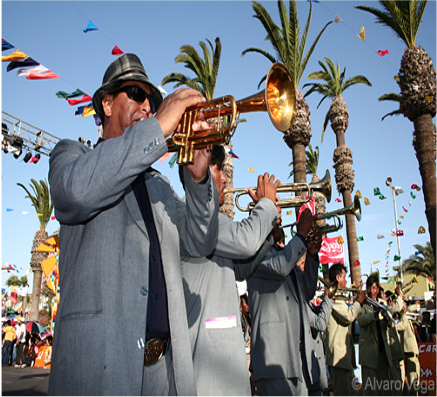} &
    \includegraphics[width=.62\columnwidth, height = 0.5\columnwidth.]{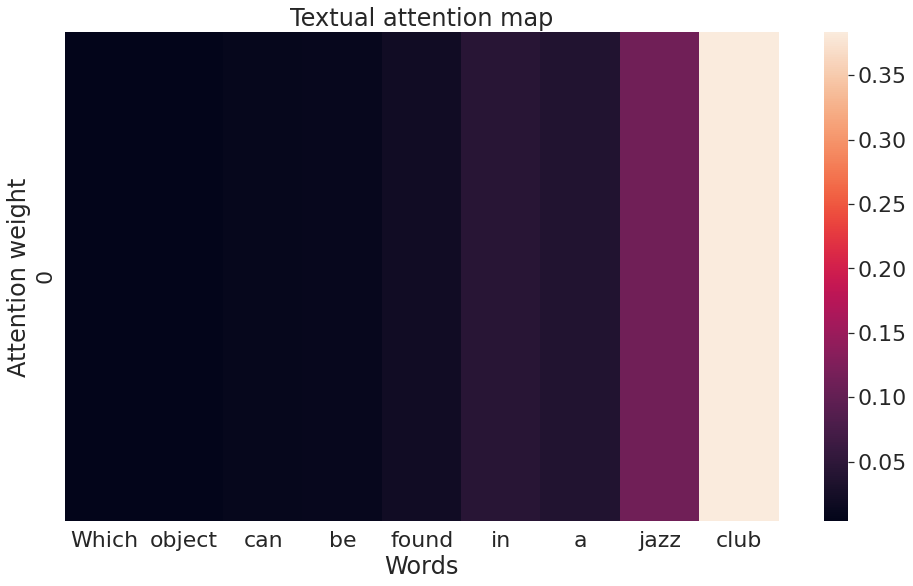} &
    \includegraphics[width=.62\columnwidth, height = 0.5\columnwidth.]{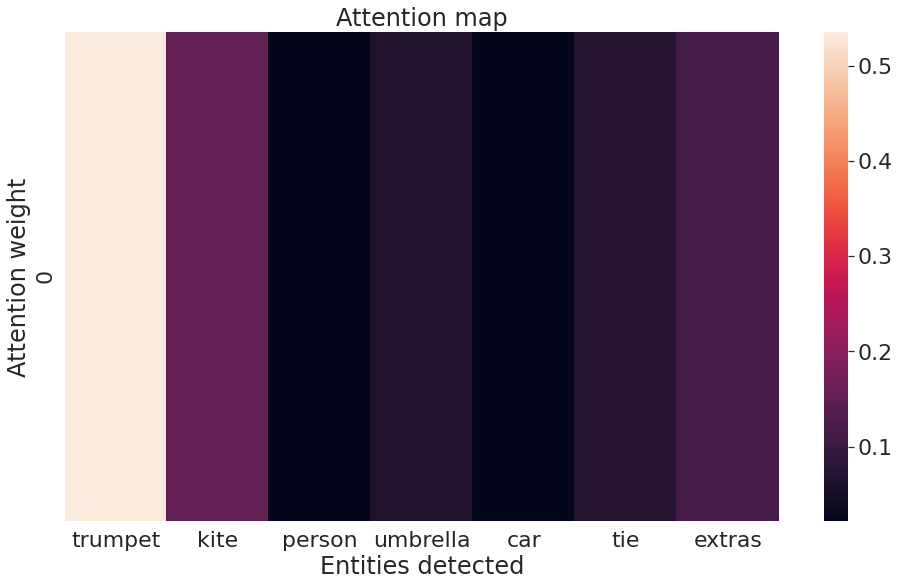} \\
      \textbf{Question 2:} Which object can be found in a jazz club?  \newline
      \textbf{SPO triple:} \{Trumpet, AtLocation, jazz club\} \newline
	  \textbf{Answer Source:} Image  \newline
	  \textbf{Answer:} Trumpet \newline
	  \textbf{Answer predicted: \textcolor{Green}{Trumpet}} 	  & & \\

    \includegraphics[width=.62\columnwidth, height = 0.5\columnwidth.]{supplements/Airplane} &
    \includegraphics[width=.62\columnwidth, height = 0.5\columnwidth.]{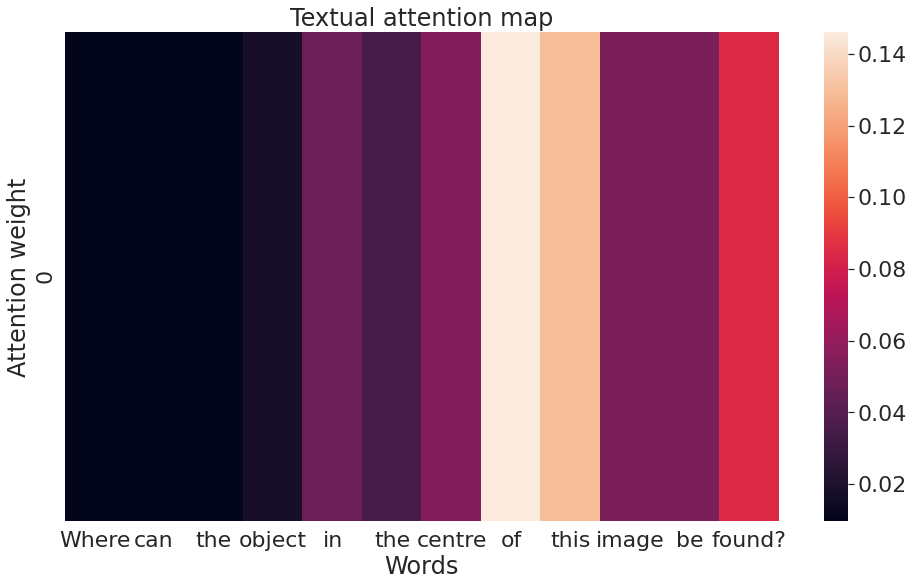} &
    \includegraphics[width=.62\columnwidth, height = 0.5\columnwidth.]{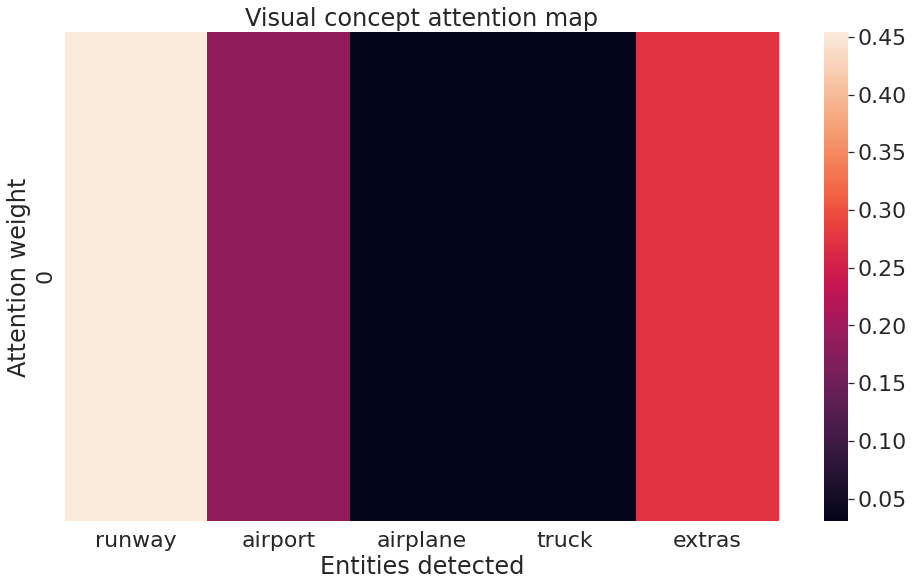} \\
      \textbf{Question 3:} Where can object in the center of the image be found?  \newline
      \textbf{SPO triple:} \{Airplane, AtLocation, airport\} \newline
	  \textbf{Answer Source:} KG  \newline
	  \textbf{Answer:} Airport \newline
	  \textbf{Answer predicted: \textcolor{Red}{Runway}} 	  & & 	  
	\end{tabular}
\caption{Visualizing CoAttention weights learnt by Seeing is Knowing}	
\label{fig:attention}
\end{figure*}

To investigate the performance of the Image-as-Knowledge module, Fig.\ref{fig:attention} depicts both the visual and textual attention maps learnt by the network. We can see that the CoAttention module is learning to pay attention to contextually relevant cues based on the input word embeddings in the question and the KG embeddings of the visual concepts. For Question 1 the textual attention correctly attends to the word 'amber and green' and the image attention correctly attends to the entity 'traffic light'. In Question 2 similarly, the textual attention attends to the word 'jazz club' while the image attention attends to the entity 'trumpet'. Likewise in Question 3, words 'centre of the image' and entity 'runway' are correctly attended to.

In cases where the answer is incorrectly predicted, the learnt attention map throws light on the reasoning of the network. 

\section{D. Choosing hyperparameters for composite score-based retrieval}
To investigate the importance of each metric in the composite score, we vary the respective coefficient $\lambda_k$ to each metric across a range of values. 

The first three rows of Table \ref{tab:composite_standard} demonstrate that the KG similarity metric individually is strongest by some margin. Considering only Jaccard similarity or GLoVe similarity is not enough to produce the accurate answer. This further highlights the crucial involvement that a KG reasoning module has in performance of any FVQA system. A standalone KG similarity metric, however, falls short of the highest achieved accuracy, owing to its relative inability to leverage lexical similarity between entity words and words in the question. (Note that the system obtained by assigning all the weight to only SiK metric is not the same as directly using SIK to retrieve the answer, since the former still uses the subgraph construction method to prune the fact search space. This is also demonstrated in the slightly different accuracy value of the two methods.)

\begin{table*}[ht]
  \centering
	\begin{tabular}{|P{1.5cm}|P{1.5cm}|P{1.5cm}|P{2.1cm}|P{2.1cm}|P{2.1cm}|P{2.1cm}|}
    \hline
\multicolumn{3}{|c|}{\textbf{Coefficient values}}& \multicolumn{2}{c|}{\textbf{Top 100 facts}}&\multicolumn{2}{c|}{\textbf{Top 500 facts}}\\
 \hline
 \textbf{$\lambda_1$} & \textbf{$\lambda_2$} & \textbf{$\lambda_3$} & \textbf{Hits @1} & \textbf{Hits @3} & \textbf{Hits @1} & \textbf{Hits @3}\\
 \hline
 \multicolumn{7}{c}{\textbf{Single similarity metric being used}}\\
    \hline
    $1$ &$0$ &$0$ & $56.05\pm0.57$ &$71.84\pm0.44$ & $55.43 \pm 0.57 $  & $69.55 \pm 0.58$ \\
    $0$ &$1$ &$0$ & $17.44\pm0.55$ &$42.98\pm0.86$ & $13.83 \pm 0.57$  & $34.64 \pm 1.00$ \\
    $0$ &$0$ &$1$ & $13.38\pm0.75$ &$30.67\pm0.80$ & $11.75 \pm 0.62$  & $28.82 \pm 0.75$ \\
   \hline

 \multicolumn{7}{c}{\textbf{Equal importance to all similarity metrics}}   \\
 \hline
    $0.34$ &$0.33$ &$0.33$ & $60.73\pm0.82$ &$77.03\pm0.46$ & $60.63 \pm 0.73$  & $76.9 \pm 0.51$ \\
    $0.4$ &$0.3$ &$0.3$ & $60.53\pm0.94$ &$\mathbf{77.14\pm0.50}$ & $60.33 \pm0.92 $  & $77.01 \pm 0.55$ \\

    $0.5$ &$0.5$ &$0$ & $\mathbf{60.82\pm0.98}$ &$75.83\pm0.42$ & $59.96 \pm 0.89$  & $74.48 \pm 0.47$ \\
    $0.5$ &$0.25$ &$0.25$ & $59.96\pm0.72$ &$76.73\pm0.37$ & $ 59.57 \pm 0.68$  & $ 76.31\pm0.57 $ \\    
    $0.5$ &$0$ &$0.5$ & $54.59\pm71.76$ &$71.76\pm0.46$ & $52.52 \pm0.74 $  & $71.03 \pm 0.68$\\
    $0$ &$0.5$ &$0.5$ & $40.18\pm1.13$ &$61.62\pm0.47$ & $39.76 \pm1.08 $  & $60.84 \pm 0.67 $ \\
	\hline        
 \multicolumn{7}{c}{\textbf{Highest importance to KG similarity}}   \\
 \hline
    $0.6$ &$0.2$ &$0.2$ & $58.89\pm0.54$ &$75.89\pm0.44$ & $58.5 \pm 0.51$  & $75.1 \pm 0.658$ \\ 
    $0.7$ &$0.15$ &$0.15$ & $58.05\pm0.64$ &$74.87\pm0.49$ & $57.6 \pm 0.5$  & $73.84\pm0.49$  \\
    $0.8$ &$0.1$ &$0.1$ & $57.23\pm0.54$ &$73.9\pm0.56$ & $56.73 \pm0.49 $  & $ 72.41\pm0.40 $ \\    
    \hline
  \end{tabular}
  \caption{FVQA performance for different convex combinations of $\lambda_k$ in the standard answer prediction task. KG Embedding algorithm used is ERMLP and image representation is Image-as-Knowledge. $\lambda_1$ is the coefficient of KG similarity, $\lambda_2$ is the coefficient of Jaccard similarity and $\lambda_3$ is the coefficient for GLoVE similarity}
  \label{tab:composite_standard}
\end{table*}

\begin{table*}
  \centering
	\begin{tabular}{|P{1.5cm}|P{1.5cm}|P{1.5cm}|P{2.1cm}|P{2.1cm}|P{2.1cm}|P{2.1cm}|}
    \hline
\multicolumn{3}{|c|}{\textbf{Coefficient values}}& \multicolumn{2}{c|}{\textbf{Only QA facts missing}}&\multicolumn{2}{c|}{\textbf{50\% KG missing}}\\
 \hline
 \textbf{$\lambda_1$} & \textbf{$\lambda_2$} & \textbf{$\lambda_3$} & \textbf{Hits @1} & \textbf{Hits @3} & \textbf{Hits @1} & \textbf{Hits @3}\\
 \hline
 \multicolumn{7}{c}{\textbf{Single similarity metric being used}}\\
    \hline
    $1$ &$0$ &$0$ & $55.23\pm0.44$&$70.56\pm0.41$&$55.20\pm0.77$&$70.47\pm0.36$ \\
    $0$ &$1$ &$0$ & $9.91\pm0.58$&$27.44\pm1.09$&$9.92\pm0.61$&$27.48\pm1.06$ \\
    $0$ &$0$ &$1$ & $12.42\pm0.72$&$29.43\pm0.72$&$12.44\pm0.76$&$29.47\pm0.72$ \\
   \hline

 \multicolumn{7}{c}{\textbf{Equal importance to all similarity metrics}}   \\
 \hline
    $0.34$ &$0.33$ &$0.33$ &$55.13\pm0.98$&$73.04\pm0.54$ & $55.13\pm1.09$&$72.95\pm0.38$\\
    $0.4$ &$0.3$ &$0.3$ &$56.35\pm1.08$&$73.56\pm0.58$ & $56.38\pm1.08$&$73.47\pm0.44$\\

    $0.5$ &$0.5$ &$0$ &$54.78\pm0.78$&$70.53\pm0.56$ & $54.79\pm0.98$&$70.48\pm0.33$\\
    $0.5$ &$0.25$ &$0.25$ &$\mathbf{56.80\pm0.82}$&$\mathbf{73.62\pm0.42}$ & $\mathbf{56.81\pm0.84}$&$\mathbf{73.54\pm0.25}$\\    
    $0.5$ &$0$ &$0.5$ &$53.68\pm0.82$&$70.77\pm0.67$& $53.63\pm0.87$&$70.67\pm0.35$\\
    $0$ &$0.5$ &$0.5$ &$26.8\pm0.90$&$49.24\pm0.62$& $26.79\pm0.85$&$49.21\pm0.63$ \\
	\hline        
 \multicolumn{7}{c}{\textbf{Highest importance to KG similarity}}   \\
 \hline
    $0.6$ &$0.2$ &$0.2$ &$56.46\pm0.51$&$73.36\pm0.51$& $56.45\pm0.63$&$73.31\pm0.22$ \\ 
    $0.7$ &$0.15$ &$0.15$ &$56.23\pm0.36$&$72.7\pm0.39$  & $56.17\pm0.66$&$72.63\pm0.34$\\
    $0.8$ &$0.1$ &$0.1$ &$55.88\pm0.25$&$72.01\pm0.46$ & $55.85\pm0.65$&$71.91\pm0.31$\\    
    \hline
  \end{tabular}
  \caption{FVQA performance for different convex combinations of $\lambda_k$ in the incomplete KG FVQA task for top-100 facts retrieved. KG Embedding algorithm used is ERMLP and image representation is Image-as-Knowledge.$\lambda_1$ is the coefficient of KG similarity, $\lambda_2$ is the coefficient of Jaccard similarity and $\lambda_3$ is the coefficient for GLoVE similarity}
  \label{tab:composite_missing}
\end{table*}
\begin{table*}[t]
  \centering
  \begin{tabular}{|P{3.5cm}|p{1.5cm}|P{1.5cm}|P{1.5cm}|P{1.5cm}|P{1.5cm}|P{1.5cm}|} 
    \hline
    \textbf{Method} & \textbf{Sampling} & \textbf{MR} & \textbf{MRR} & \textbf{Hits@1} & \textbf{Hits@3}& \textbf{Hits@10}\\
    \hline
    TransE & adversarial & $14692$  & $0.1035$ & $0.05$ & $0.13$ & $0.19$\\
    TransE & uniform  & $13736$  & $0.0967$ & $0.04$ & $0.112$ & $0.194$\\
    RotatE  & adversarial  & $17660$  & ${\bf 0.173}$ & ${\bf 0.14}$ & ${\bf 0.18}$ & ${\bf 0.22}$ \\
    RotatE & uniform & $18468$  & $0.144$ & $0.122$ & $0.152$ & $0.179$ \\
    ERMLP & adversarial & $\textbf{11194}$  & $0.156$ & $0.132$ & $0.152$ & $0.197$ \\
    ERMLP & uniform & $14907$  & $0.122$ & $0.09$ & $0.128$ & $0.173$ \\    
    \hline
  \end{tabular}
  \caption{KG Embedding accuracy for standard retrieval task}
  \label{tab:kge_results_app}
\end{table*}
\begin{table*}[t]
  \centering
  \begin{tabular}{|P{8cm}|P{3cm}|P{3cm}|} 
    \hline
    \textbf{Technique} & \textbf{Hits@1} & \textbf{Hits@3}\\
    \hline
    TransE, IaK & $38.64 \pm 1.51$ & $52.58 \pm 1.93$ \\
    TransE – Adv, IaK & $40.37 \pm 0.87$ & $53.1 \pm 1.12$ \\
    RotatE, IaK & $45.59 \pm 0.52$ &  $56.55 \pm 1.18$\\
    RotatE – Adv, IaK & $47.92 \pm 1.15$ & $61.3 \pm 0.97$ \\
    ERMLP, IaK & $53.16 \pm 0.79$ & $63.9 \pm .63$\\
    ERMLP -Adv, IaK  & $\mathbf{54.38 \pm 0.94}$ & $\mathbf{65.76\pm 0.5}$\\    
    \hline
  \end{tabular}
  \caption{FVQA accuracy for standard retrieval task (IaK - Image as Knowledge, Adv - self-adversarial negative sampling}
  \label{tab:fvqa_results_app}  
\end{table*}

The results in the next section in the table demonstrates that lexical and distributional semantic features indeed hold complementary information to that of KG embeddings. To remedy the trend of SiK ignoring lexical similarity, roughly equal importance is assigned to the Jaccard and GLoVE similarity metrics. Interestingly, the convex combination works best when each distance metric is weighted roughly equally. Our experiments indicate that, the majority of the correct answers are produced due to the similarity score arising from SiK, but an equal weighting is required so as to nudge the correct answer ahead of the incorrect ones. 

Lastly, we observe that weighting the SiK metric more than 0.6 starts to yield lesser and lesser benefits, and the performance monotonically decreases. 

In both the incomplete KG settings (Table \ref{tab:composite_missing}), we observe that pruning the fact search space still improves accuracy over the standalone networks. But the contribution of the textual similarity metrics becomes almost negligible compared to its impact when required facts are present in the network. 

\section{E. Performance of all KG Embedding algorithms}
ERMLP has better mean-rank, while RotatE performs better in all other standard metrics for KG embeddings. (Table~\ref{tab:kge_results_app}). However, ERMLP-based embeddings are more successful for FVQA (Table~\ref{tab:fvqa_results_app}). Apparently ERMLP adds more capacity to the embedding model, which also helps the downstream task of FVQA.

Using self-adversarial negative sampling is beneficial to all KG Embedding algorithms. The FVQA accuracy improves by 1.67\%, 2.47 \%, and 1.25\% respectively for TransE, RotatE, and ERMLP. 
. 

\end{document}